\documentclass{elsarticle}
\usepackage{lipsum}
\makeatletter
\def\ps@pprintTitle{%
 \let\@oddhead\@empty
 \let\@evenhead\@empty
 \def\@oddfoot{}%
 \let\@evenfoot\@oddfoot}
\makeatother

\usepackage{lineno}

\journal{Journal Name}

\begin{document}

\begin{frontmatter}


\title{A GA-based feature selection of the EEG signals by classification evaluation: Application in BCI systems}

\author[1]{\small Samira Vafay Eslahi}
\author[2]{\small Nader Jafarnia Dabanloo}
\author[3]{\small Keivan Maghooli}
\address[1]{\footnotesize Department of Electrical and Computer Engineering, Texas A\&M University, College Station, TX, USA\\

}
\address[2]{\footnotesize Department of Biomedical Engineering, Science and Research Branch, Islamic Azad University, Tehran, Iran}
\address[3]{Department of Biomedical Engineering, Science and Research Branch, Islamic Azad University, Tehran, Iran}
\begin{abstract}
In electroencephalogram (EEG) signal processing, finding the appropriate information from a  dataset has been a big challenge for successful signal classification. The feature selection methods make it possible to solve this problem; however, the method selection is still under investigation to find out which feature can perform the best to extract the most proper features of the signal to improve the classification performance. In this study, we use the genetic algorithm (GA), a heuristic searching algorithm, to find the optimum combination of the feature extraction methods and the classifiers, in the brain-computer interface (BCI) applications. A BCI system can be practical if and only if it performs with high accuracy and high speed alongside each other. In the proposed method, GA performs as a searching engine to find the best combination of the features and classifications. The features used here are Katz, Higuchi, Petrosian, Sevcik, and box-counting dimension (BCD) feature extraction methods. These features are applied to the wavelet subbands and are classified with four classifiers such as adaptive neuro-fuzzy inference system (ANFIS), fuzzy k-nearest neighbors  (FKNN), support vector machine (SVM) and linear discriminant analysis (LDA). Due to the huge number of features, the GA optimization is used to find the features with the optimum fitness value (FV). Results reveal that Katz fractal feature estimation method with LDA classification has the best FV. Consequently, due to the low computation time of the first Daubechies wavelet transformation in comparison to the original signal, the final selected methods contain the fractal features of the first coefficient of the detail subbands.
\end{abstract}


\end{frontmatter}

\section{Introduction}
\label{S:1}
BCI technology provides a direct communication between people and computer. It monitors the EEG signals and consequently the brain activity for detecting the related tasks via signal processing algorithms to enable the communication\cite{c0}. It has the potential to enable the physically disabled patients to perform the activities with a higher quality and can increase their productivity. The most common motor imagery tasks are hand \cite{c8}, feet \cite{c0}, and tongue \cite{c9} movements. BCI systems can translate these tasks to the computer commands. The signals from the motor cortex should be classified to be converted to the computer language. One of the most important criteria for a good classification is the accuracy. As the BCI applications demand, the computation time should be very low to convert the motor imagery tasks as fast as possible to the computer commands. The trade-off between accuracy and computation time has been a big challenge, which desires the EEG optimization methods. To improve the classification accuracy classifiers and features play important roles. Different studies investigated the usage of the GA to find the most efficient feature sets. For instance, continuous and binary GA optimization has been proposed to characterize the patient's epilepsy risk level \cite{c25}. The other method decomposed the normal and epileptic seizer EEG signals into different frequencies with 4th level of the wavelet and used the approximation entropy of the decomposition nodes as the feature sets for classification \cite{c26}.  They used GA as the optimization method to find the risk of epilepsy. The other GA based optimization proposed in 2010. They decomposed the EEG signals into five subband components. The features they considered in that method were the nonlinear parameters that were classified by support vector machine (SVM)  with linear kernel function (SVML) and radial basis function kernel function (SVMRBF) classifiers \cite{c27}. The other feature selection method has been proposed with GA that was evaluated by SVM \cite{c28}. A proposed method showed that with GA optimization, the performance of the neural network would be improved \cite{c29}.
Fractal dimension estimation is a statistical measurement indicating the complexity of an object or a quantity that is self-similar over some regions of space or time interval. The fractal feature estimation has been successfully used in various domains to characterize the objects and quantities, but its usage in motor imagery tasks in BCI applications is still under the investigation and has been widely used in the last decades \cite{c12}\cite{c13}. The biggest challenge according to the BCI is to extract features for acceptable speed and accuracy. Numerous feature extraction methods use fractal dimension estimation to extract geometrical features from signals. The most famous methods that calculate fractal dimension of a signal are Katz\cite{c23}, Higuchi\cite{c14}, Petrosian \cite{c15}, Sevcik\cite{c11}, and BCD feature extraction methods. The combination of Katz and Higuchi fractal dimensions with Fuzzy k-nearest neighbors (KNN) \cite{c18}, SVM \cite{c19}, and linear discriminant analysis (LDA) \cite{c20} classifiers has been purposed \cite{c16}. They showed that combination of the Katz method with the FKNN classifier has the best performance based on the time and accuracy in comparison to the other methods.  Then they modified the best methodology of the combination of these features and classifiers called time-dependent fractal dimension (TDFD) \cite{c21}, differential fractal dimension (DFD) \cite{c22}, differential signals (DS) \cite{c16}.
In this study, we investigate the combination of different classifiers such as ANFIS, FKNN, SVM, and LDA with different feature extraction methods such as Katz, Higuchi, Petrosian, Sevcik, box-counting dimension, on first the original data and second on the detail and approximation sub-bands of the wavelet transformation. The classification is based on the right or left-hand imagery movements and evaluates the fractal features on both the subbands and the original signal. Finally, the computation time and accuracy are optimized by GA to find the best method with the highest performance.

\section{Materials and Methods}
\label{S:2}

\subsection{Dataset}
The data is downloaded from BCI Competition 2003, Dataset III \cite{c30}. A female subject sat in a chair with keeping her arms relaxed. By means of imagery left or right-hand movements the task was performed, and a feedback bar controlled the process. The EEG signal was recorded with a sampling rate of 128HZ from three channels at standard positions of the 10-20 international system (C3, Cz, and C4). The signals were filtered between 0.5 and 30 HZ. Each run had 40 trials and each trial was nine second long. The imagery was done for about six seconds with a cue indicating that the task was presented. The task contains 280 trials. In this study, 9 fold is using for trains and 1 fold is used for test the classifiers' performances.

\subsection{Optimization}
A genetic algorithm is a searching engine, which is based on natural selection and genetics that was proposed by Holland \cite{c5}. GA has four steps as follows:
Step 1: Initializing and fitness calculation: First step in the Genetic algorithm is generating a random population of chromosomes. These chromosomes encode the solution candidates called individuals. For optimizing a problem, GA calculates the fitness value of each of the chromosomes.
Step 2: selection: During each iteration, the chromosomes that have higher fitness values will be eliminated, and consequently the chromosomes with lower fitness values will be selected as the desired parents to produce children for the next generation.
Step 3: reproduction and Mutation: After reproducing the population, mutation technique is used. This procedure prevents the algorithm to fall in the local minimums. It flips some Bits of chromosomes randomly to change them in that generation.
Step 4: Termination. This process is repeated until the condition is met. 
\begin{figure*}[!ht]
  \centering
 \includegraphics[width=\textwidth]{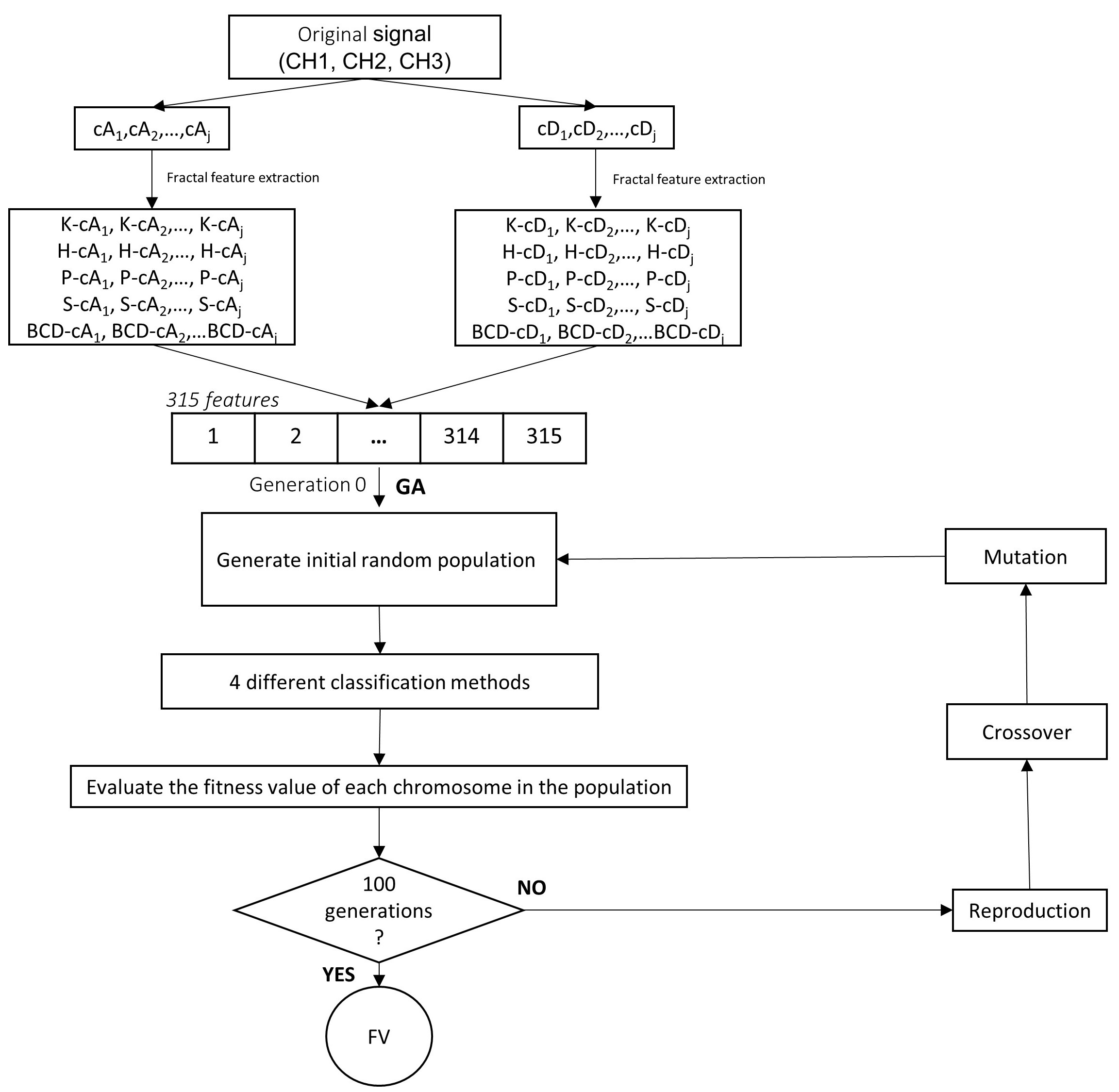}
  \caption{\footnotesize The flowchart of the proposed method. The fractal features are extracted from both wavelet coefficients and the original signal. The combination of these features results in a large number of features in the feature space. In addition. four classifiers are used for the classification. In each iteration, the winners will be transferred to the next generation, and the losers will be eliminated from the rest of the procedures.  }\label{figure 1}
\end{figure*}
Figure 1 shows the process of the proposed method. With the discrete wavelet transformation, the signal is decomposed into high and low frequencies with the dB2 of Daubechies, the most popular family of the wavelet transformation. This procedure continues to level 10 to find 10 details and 10 approximation coefficients. In this stage, EEG digital signal is decomposed into detail and approximation subbands. The coefficients of the details and approximation subbands are calculated up to 10 levels as the rest does not have enough information about the signal. In consequence, we calculated fractal dimensions of these subbands and obtained numerous feature sets. In the trade-off between the computation time and accuracy, we evaluate the performance of the methods with the minimum computation time and maximum classification accuracy. The evaluation is done by GA and with a defined fitness function. The considered fitness function is the ratio of the time to accuracy that a high FV shows a poor performance and a low FV shows an optimum performance. GA optimization works as follow:
The initial population consists of N parents and consequently N chromosomes. Each chromosome is a string of Genes and has the length of L. Each gene is a binary allele that is a selected feature from the feature space. Calculating the fractal dimensions of the approximation and detail subbands with five feature estimation methods make the feature space to be comprised of 100 features. Ten approximation and ten detail subbands with five methods of fractal estimation make the feature sets contains 100 features. On the other hand, three channels producing these 100 features would result 300 features with different information. We also add more 15 features that obtained from applying five fractal dimension calculations on original signal from the 3 channels. With these combinations, the feature space consists of 315 different features with different information about the signal. 
The feature space is comprised of 315 different features, which can be defined as the initial population contains parents with different genes in one chromosome in GA optimization procedure. The number of combined features for the best classification results depends on the accuracy and computation time. Increasing the number of features combined can increase the computation time, which is not favorable while increasing the number of features can improve the accuracy of the classification. The number of different combined features can be defined by the best fitness value calculated by GA. As the number of features is 315, we defined the initial population with choosing a couple of parents that each of them has a chromosome contains 315 genes. Each chromosome is a binary string that contains only zeros and ones. Therefore each gene included in the binary string can be assigned to two numbers, one and zero, in which one represents the selection of one specific feature, and zero means that the same feature is not selected (Fig 2).
\begin{figure}[!ht]
  \centering
  \includegraphics[scale = 0.59]{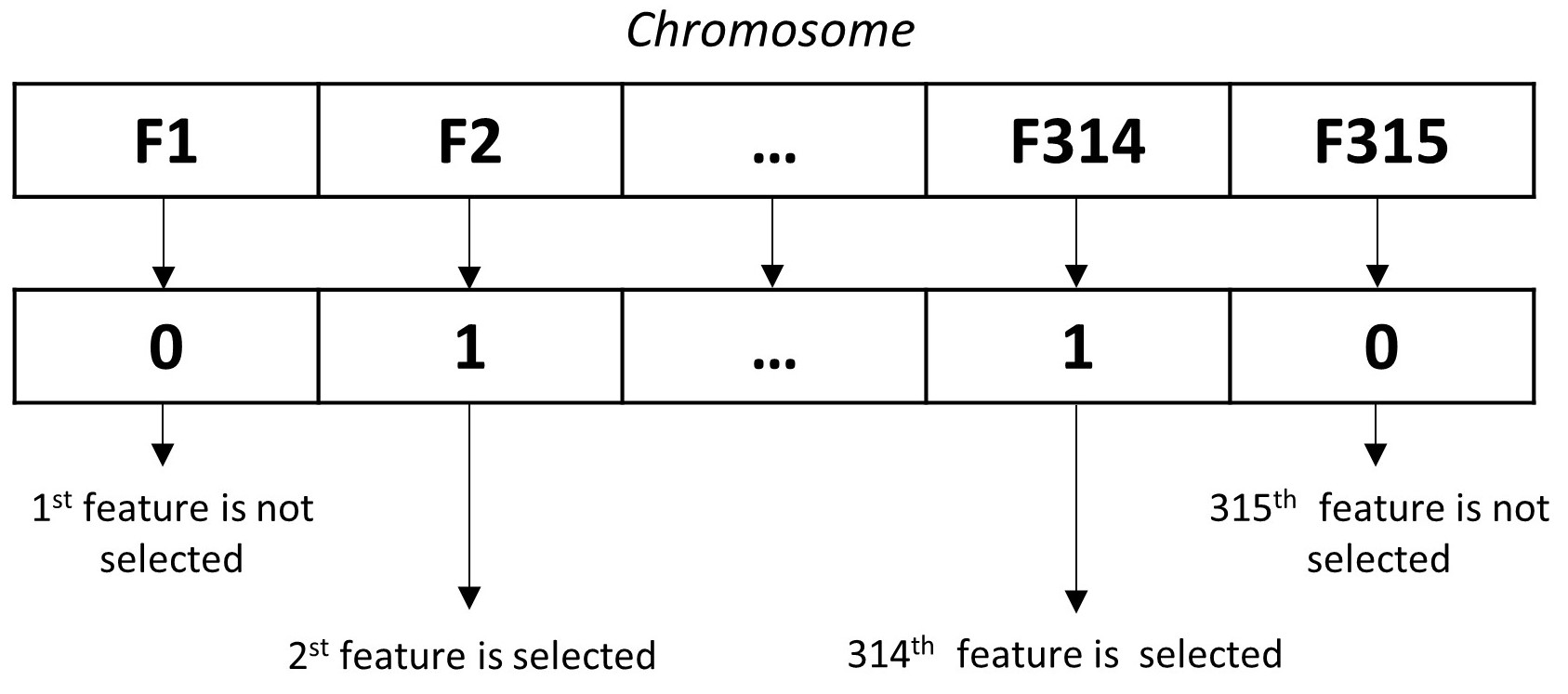}\label{fig:figure 2}
  \caption{\footnotesize A single chromosome of a parent. Each of the genes in the chromosome string will be zero or one. Each feature contains its original place in the new chromosome. If the gene shows zero, it means that that feature from the same place is not selected, and if the gene shows 1, it means that the feature from that specific gene is selected. Combination of the genes will be a new feature that should be classified.}
\end{figure}
In each generation, two parents are selected among number of parents, and the five classifiers classify the feature combination of each of these parents separately. The fitness value of each of these classifications is calculated and evaluated. The termination of the GA is defined as meeting the 100 generations that means when the generation passed the 100th generation the algorithm stops the optimization. In fact, the FV of these parents shows how effective each parent and the offspring are. The procedure continues as the methods with high FV are stopped and the methods with low FV are transferred the next generation, which is producing the offspring.

The selected parents will be sent to the crossover procedure, which means half of the genes of the first parent will be combined with half of the genes of the second parent. The combination of the features included in the chromosome of each of the parents was selected that was the best among the calculated ones. Therefore, these combinations will be well enough to be transferred to the next generations. The optimization continues until the generation meets the defined number of generations. As the feature space is very large, a threshold has been defined to eliminate the high FV of the classified signal. The threshold has been defined 5\% of the maximum FV, which is the combination of all the features. This defined threshold can prevent the extra calculations. The 5\% is set from defining a limited number of calculation, which will be about 7 features in a combination here (Fig 3). 

\begin{figure}[!ht]
  \centering
  \includegraphics[scale = 0.59]{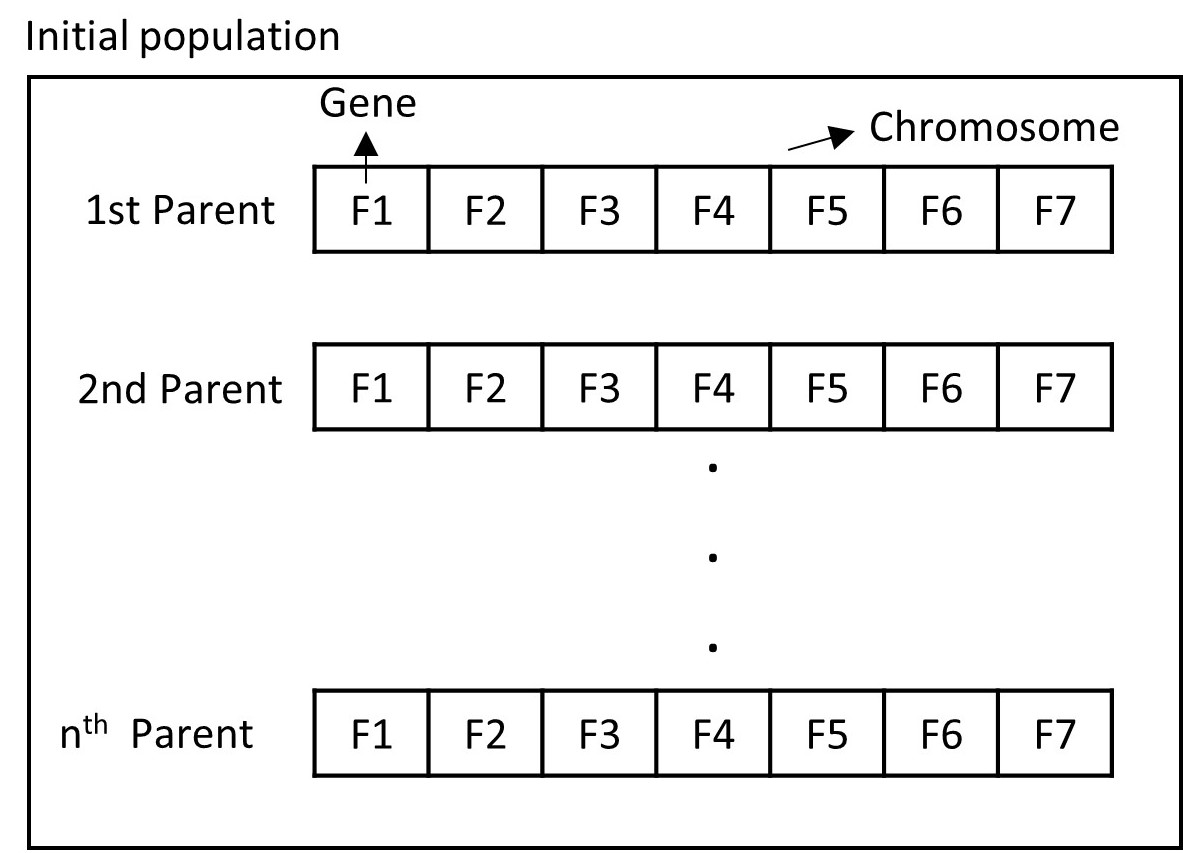}\label{fig:figure 3}
  \caption{\footnotesize The initial population. The initial population consists of 200 parents that each parent has a chromosome contains about 7 genes. Each gene represents a special feature in the feature set. For example, 011010...01101 means that the second, third, fifth, 312th, 313th and 315th have been selected that make a combination of 7 features in the chromosome. The chromosomes will be scatter crossed over to generate the next generation.}
\end{figure}

The fitness value has been defined the ratio of the computation time over accuracy, and the lower value of the function is transferred to the next generation to find the best combination of the features and the best combination of the features with classifiers. The fitness function is defined by,\\  
FV=T/A\\
where $FV$ is fitness value, $T$ is the computation time after the classification, and $A$ is the classification accuracy. In GA optimization, the selection, crossover, and mutation should be considered wisely to prevent the elimination of the combination, which are favorable. The selection is set by a selection rate and the crossover probability of 0.9 to choose the parents with lower fitness values for generating the next population. Parent with higher fitness values will be eliminated to reduce the time of optimization. On the other hand, the crossover of the selected parents is a technique that chromosomes exchange their genes with each other to produce children for the next population. Because the orders matter here, the scattering type of crossover has been considered, which means the randomly selected genes will be selected in 2 chromosomes and will be exchanged their places. The scatter crossover is considered due to the changing the order of the features combination. The scatter crossover prevents the combination of features to be repeated in the other generations. For instance, if F1-F2-F3-F4-F5-F6-F7 is one of the generated offspring and F8-F9-F10-F11-F13-F14-F15 is the other one, the next generation is not using F-1-F2-F3-F4-F13-F14-F1 in this order necessarily. In fact, the other probabilities of the lower FV have been investigated too.  The new chromosomes are children of the previous generation, or in the other words, they are the parents of the new generation. Furthermore, in the mutation process, some genes randomly, and with a rate of 0.001 will be changed. This means that a number between zero and one is produced randomly in which if the number is lower than the defined mutation rate, a random binary string is produced and one random gene in the actual chromosome will be exchanged with the newly generated chromosome. This will avoid the algorithm to fall in the local minimums. Finally, by meeting the termination criteria i.e. 100 generations, the algorithm is converged to a minimum value that is the least fitness function value in the 100 generations. In addition, the crossover of the best couple is selected from the previous generation that necessarily has not a value better than previous parents do; therefore, the minimum fitness value will have an elite of to the next generation on the condition that the elite parent has a very low value that the elitism can be applied to them.

In assigning values to GA parameters, preventing GA to trap in the local optimums is one of the most important factors that have to be considered. In addition, the divergence of future generations should be controlled to have an acceptable result in real-time signal processing. The attributed values to GA parameters are based on previous similar researches and with some little changes in these values we tried to adjust the GA parameters by examination to reach to the optimum value.

\section{Results}
\label{S:3}
All the simulations were done by MATLAB R2011b (7.13.0.564). Figure 4 shows the GA convergence to the minimum value and shows how the optimization process could find the features that with the defined FV has the minimum value.   
\begin{figure}[!ht]
  \centering
  \includegraphics[scale = 0.6]{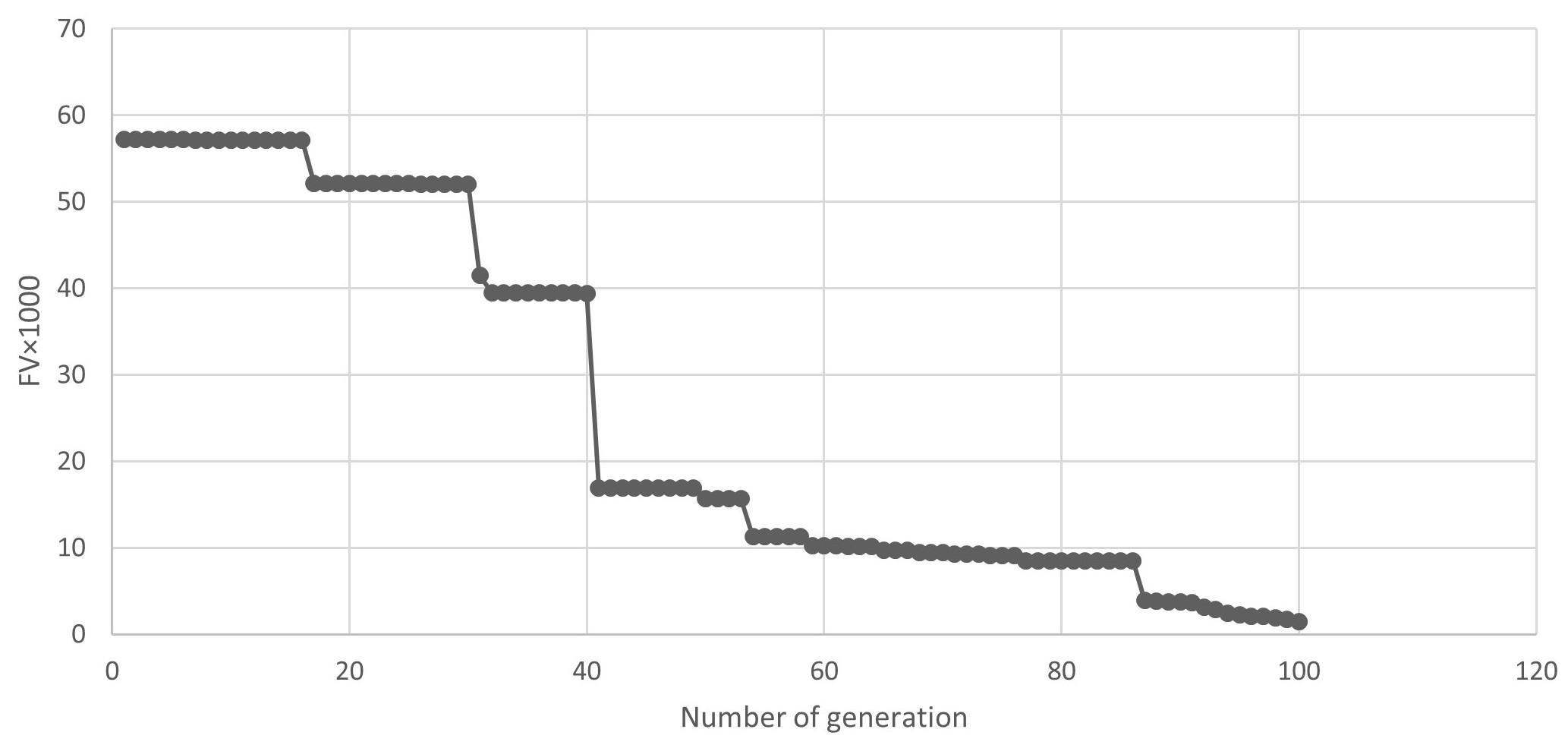}\label{fig:figure 4}
  \caption{\footnotesize GA convergence. the generations start from zero which is the initial population. The next 100 generations have been produced and optimized to be converged to the minimum value. The minimum value is the best method selected from this optimization algorithm.}
\end{figure}

\begin{table}[h]
\caption{\footnotesize GA selected methods with the lowest FV. The accuracy and the computation time are calculated.  }
\label{table 1}
\begin{center}
\begin{tabular}{|p{4cm}|p{2.2cm}|p{1.7cm}|p{1.5cm}| }
\hline
Feature(s)-classifier  & Accuracy (\%) &Time (sec) & \hspace{1mm} FV\\
\hline
Petrosian on D1-ANFIS & \hspace{8mm} 74 &  \hspace{3.2mm} 0.29 &\hspace{0.4mm} 0.0039\\
\hline
B.C.D-SVM	& \hspace{8mm} 71	& \hspace{3.2mm} 0.27&\hspace{0.4mm} 0.0038\\
\hline
DS-ANFIS&	\hspace{8mm} 67	&  \hspace{3.2mm} 0.25&	\hspace{0.4mm} 0.0037\\
\hline
DS-FKNN&	\hspace{8mm} 70	&  \hspace{3.2mm} 0.26	&\hspace{0.4mm} 0.0037\\
\hline
B.C.D-FKNN	& \hspace{8mm} 68&	 \hspace{3.2mm} 0.25&\hspace{0.4mm}	0.0037\\
\hline
DS-SVM	& \hspace{8mm} 73&	 \hspace{3.2mm} 0.23&\hspace{0.4mm}	0.0031\\
\hline
B.C.D on D1-FKNN&	\hspace{8mm} 77	&  \hspace{3.2mm} 0.22	&\hspace{0.4mm} 0.0029\\
\hline
Katz on D1-LDA&	\hspace{8mm} 79	&  \hspace{3.2mm} 0.19	&\hspace{0.4mm} 0.0024\\
\hline
Katz on D1-SVM&	\hspace{8mm} 75	&  \hspace{3.2mm} 0.17	&\hspace{0.4mm} 0.0023\\
\hline
Katz-FKNN&	\hspace{8mm} 83	&  \hspace{3.2mm} 0.17	&\hspace{0.4mm} 0.0020\\
\hline
B.C.D on D1-LDA	& \hspace{8mm} 69	& \hspace{3.2mm} 0.14&\hspace{0.4mm} 0.0020\\
\hline
Katz on D1-FKNN&	\hspace{8mm} 80&	 \hspace{3.2mm} 0.14	&\hspace{0.4mm} 0.0017\\
\hline
Katz-LDA&	\hspace{8mm} 84	&  \hspace{3.2mm} 0.12	&\hspace{0.4mm} 0.0014\\
\hline
\end{tabular}
\end{center}
\end{table}
The results showed that the feature combination has the best result whenever the combination did not exceed of more than two features in combination. The computation time will be increased when the number of features that cooperates in the classification and there is not a significant change in the classification. By a different fitness function obviously the results are different, and it depends on the importance of the factors involved in the final FV. The convergence of the GA shows that GA is a strong optimization algorithm that can find the optimum solution in a large feature set. The minimum fitness values that were found by the proposed method is shown in Table 1. 
In some applications, the accuracy is more important than the computation time; therefore fitness function can be changed by adding a weighting factor to the accuracy. In BCI applications, the computation time plays a significant role, as the brain signals should be transferred to the computer as fast as possible, and hands move as soon as the brain imagine the movements. The results show that the proposed method can select the best features in a large feature set and can investigate the combination of them. 

\section{Discussion}
\label{S:4}
In this paper, we optimized the combination of the fractal features and classification of these combined feature sets. This optimization is based on the extraction of the fractal features from approximation and detail sub bands acquired by wavelet transform and the original signal. In classification, we used four types of classification methods to recognize the class of motor imagery movements such as the task related to the right or left-hand movement. First, we passed the recorded signal from the high passed and low passed filters and decomposed the signal to two sub band sets as approximation and detail. Then we used five famous methods of measuring fractal dimensions such as Katz, Higuchi, Petrosian, Sevcik and Box Counting Dimension algorithms to calculate the geometry dimensions of these extracted sub bands. In this method, due to the huge number of features, we used GA to decrease the dimension of the feature space. The best accuracy was found from Katz feature estimation method that has been classified with LDA. The program is run multiple times and the results differed each time, but one of the results that reached to the minimum fitness value selected as the best results of the GA optimum solution. Katz features with the LDA classifier was the winner in all of the runs. Any of the feature sets and classification methods have their own advantages and disadvantages; however, the best method is defined by a method that has the lowest fitness value among the investigated methods and can have an acceptable performance for EEG classification. In this study, the fitness value has been calculated by the ratio of the computation time over the classification accuracy, in which it can result in different values with different considered fitness functions. The considered weighted factor for both computation time and the classification accuracy is one, but by adding a weighting factor for any independent value in the fitness function can make it more reliable. The proposed method is evaluated by 10-fold cross-validation. As mentioned the data consists of 280 trials that nine fold was applied in the training of the classifications and 1-fold for testing the performance of the data.

\section{Conclusion}
\label{S:5}
Genetic algorithm plays an important role as a feature selection algorithm to choose optimum features among numerous choices. Producing the generation of children from parents with the best fitness values can cause the targeted calculation to optimize time and additional computations. In this paper, we applied genetic algorithm as a feature selection method to choose the optimum fitness values. These values are based on features that are the combination of the fractal dimensions of original signal and subbands obtained from wavelet transform. We could reach the results that have the minimum fitness values among compared methods. Results showed that the proposed method could perform as a strong algorithm for feature selection. In other words, this searching algorithm with consideration of the computation time and classification accuracy can solve the problem of minimizing the fitness value to optimize the algorithm and extract features that are more appropriate for motor imagery movements of the right and left hand. In the ANFIS implementation, the computation time has been calculated only on testing data. As each of the individual person's EEG signals should be trained by itself, training of the other signals may not be reliable for all of the other individuals. In a certain case, it is reliable to train the data once the person imagined a task and the EEG data was acquired. The reason is that a person's EEG signals can remain the same any time that the specific task is imagined. Therefore, in this study, the hypothesis is based on any individual who repeats any specific task repeatedly. It is in a great interest to find an algorithm for training the signals of different participants and test that with a random individual. In addition, this research would open a way to optimize some parameters that are not considered in the classification training. These parameters can lead to different results in different situations. Furthermore, there are some other parameters in fractal dimensions, which are better to be optimized. We can use the genetic algorithm to optimize these parameters to reach a minimum defined nonlinear ratio of time to accuracy. Defining a nonlinear fitness value would be more realistic in the commercial BCI systems.\\

References

\end{document}